\documentclass[conference]{IEEEtran}
\IEEEoverridecommandlockouts

\usepackage{cite}
\usepackage{amsmath,amssymb,amsfonts}
\usepackage{algorithmic}
\usepackage{graphicx}
\usepackage{textcomp}
\usepackage{xcolor}
\usepackage{url}

\usepackage{booktabs}
\usepackage{array}
\usepackage{tabularx}
\usepackage{float}
\usepackage{placeins}
\usepackage{enumitem}

\usepackage{listings}
\usepackage{inconsolata}

\usepackage{times}
\usepackage{latexsym}
\usepackage[T1]{fontenc}
\usepackage[utf8]{inputenc}
\usepackage{microtype}

\usepackage{tikz}
\usetikzlibrary{arrows.meta, positioning, calc, fit, backgrounds, shadows.blur}

\def\BibTeX{{\rm B\kern-.05em{\sc i\kern-.025em b}\kern-.08em
    T\kern-.1667em\lower.7ex\hbox{E}\kern-.125emX}}

\begin{document}

\title{Internalizing Tool Knowledge in Small Language Models via QLoRA Fine-Tuning}

\author{
\IEEEauthorblockN{
Yuval Shemla\IEEEauthorrefmark{1},
Ayal Yakobe\IEEEauthorrefmark{2},
Tanmay Agarwal\IEEEauthorrefmark{2},
Dhaval Patel\IEEEauthorrefmark{3},
Kaoutar El Maghraoui\IEEEauthorrefmark{2}
}
\IEEEauthorblockA{\IEEEauthorrefmark{1}Columbia School of General Studies, Columbia University, NY, USA}
\IEEEauthorblockA{\IEEEauthorrefmark{2}Columbia Engineering, Columbia University, NY, USA}
\IEEEauthorblockA{\IEEEauthorrefmark{3}IBM Research, NY, USA}
\IEEEauthorblockA{
ys3571@columbia.edu,
amy2127@columbia.edu,
ta2830@columbia.edu,
pateldha@us.ibm.com,
kaoutar@cs.columbia.edu
}
}

\maketitle

\begin{abstract}
Tool-using language agents commonly include full tool schemas in every prompt, even when the available tools are fixed across many queries. This repeated schema context increases input length and can make small models unreliable planners. We study whether small language models can internalize a fixed tool catalog through supervised parameter-efficient fine-tuning, enabling structured tool planning without explicit tool descriptions at inference time.

Using AssetOpsBench, an industrial asset-operations benchmark with MCP-style tools, we fine-tune Gemma 4 E4B and Qwen3-4B with 8-bit QLoRA on approximately 1,700 tool-use examples. Under description-free inference, the fine-tuned models outperform an unfine-tuned schema-informed baseline that receives the full tool catalog, while reducing prompt length by 94.7\%. The best Gemma run achieves an AT-F1 of 0.65 and an overall LLM-judge score of 3.88, compared with 0.47 and 2.88 for the informed baseline. 

These results suggest that, for fixed tool catalogs, supervised QLoRA fine-tuning can shift tool knowledge from prompt context into adapter weights, reducing repeated schema overhead while maintaining or improving tool-planning quality.
\end{abstract}

\section{Introduction}

Large language models (LLMs) are increasingly used as planning components in tool-using agents, where they decompose user requests, select tools, construct arguments, and coordinate multi-step workflows. In many current tool-use pipelines, including MCP-style systems \cite{b1}, the model receives a serialized tool catalog in the prompt so that it can infer which tools exist and how they should be called. This schema-informed approach is flexible: new tools can be added by updating the prompt. However, it also repeats the same catalog across queries, creating substantial input overhead.

This paper studies a narrower but practically important regime: \emph{fixed-catalog tool planning}. In many enterprise and industrial deployments, the available tool catalog changes slowly relative to the number of user queries. In such settings, repeatedly prompting with full tool descriptions may be unnecessary. We ask whether this stable catalog knowledge can instead be amortized into a small model through supervised fine-tuning, allowing the model to generate structured tool-use plans without receiving tool descriptions at inference time.

This setting targets two related inefficiencies. First, small language models often require extensive tool-context prompting to produce valid plans. Second, repeated schema prompting can dominate the input budget. In our AssetOpsBench setting \cite{b2}, the serialized tool catalog accounts for approximately 2,200 of the 2,400 prompt tokens, while the user query itself is short. Description-free inference removes this catalog section entirely, reducing the prompt to approximately 128 tokens and asking the fine-tuned model to recover server routing, tool selection, argument keys, and dependency structure from its learned parameters.

\subsection{Contributions}

We propose and evaluate \emph{description-free} tool planning: a fixed-catalog setting in which the model must generate structured tool-use plans without receiving tool descriptions at inference time. Our contributions are:

\begin{enumerate}
\item We formalize \emph{static-catalog description-free tool planning}, where a model must select MCP servers, tools, arguments, and dependencies for a fixed catalog while omitting the serialized tool catalog from the inference prompt.
\item We develop a catalog-grounded supervised fine-tuning protocol in which a teacher model has access to full tool schemas during data construction, while the student model is trained and evaluated with the tool catalog removed.
\item We show that 8-bit QLoRA fine-tuning on approximately 1,700 examples enables $\sim$4B parameter models to outperform an unfine-tuned schema-informed baseline on AssetOpsBench, despite receiving no tool descriptions at inference time.
\item We evaluate whether the resulting specialization preserves broader model capability, showing that LoRA rank induces a quality--retention trade-off between tool-planning performance and general multiple-choice accuracy.
\end{enumerate}

\section{Related Work}

\subsection{Tool-Use and Function-Calling Evaluation}

Tool-use evaluation has evolved from measuring whether models can call individual APIs to testing tool selection, argument construction, dependency tracking, and multi-step execution. API-Bank \cite{b25} introduced a runnable benchmark for tool-augmented LLMs with 73 API tools. BFCL \cite{b24} evaluates function calling across serial and parallel invocations using AST-based evaluation. MCP-Bench \cite{b19} extends this to MCP-style agents, evaluating multi-step tasks that require cross-tool coordination across live MCP servers. Our work does not propose a new benchmark; instead, we study a stricter inference condition within a fixed-domain benchmark: the model must plan without receiving tool descriptions at inference time.

\subsection{Schema-Informed and Retrieval-Augmented Tool Learning}

A dominant approach to tool use is to expose tool information at inference time through prompt schemas or retrieved documentation. Gorilla \cite{b9} fine-tunes models for API invocation with retrieval-aware training. ToolLLM \cite{b10} constructs ToolBench over 16,464 real-world APIs with a neural API retriever. ToolACE \cite{b26} shows that high-quality synthetic function-calling data with automatic verification can train strong small models. These approaches are well suited to broad or changing API ecosystems. Our setting makes the opposite trade-off: we assume a fixed, mature tool catalog and ask whether its schema information can be amortized into model parameters.

\subsection{Tool and Skill Internalization}

Recent work has begun to move tool or skill knowledge from prompt context into model parameters. ToolGen \cite{b22} represents each tool as a unique vocabulary token, integrating tool knowledge into the language model's generation process and enabling unified tool retrieval and calling across over 47,000 tools. SKILL0 \cite{b23} studies skill internalization for embodied agents by providing skill context during RL training and progressively withdrawing it until the agent operates without runtime skill retrieval. Our work is motivated by the same observation, prompt-time tool context is expensive, but uses a different approach: ordinary supervised fine-tuning with QLoRA on a fixed industrial catalog, without special tool tokens, vocabulary expansion, or reinforcement learning.

\subsection{Prompt Compression and Schema Removal}

Prompt-compression methods such as LLMLingua \cite{b11} reduce inference cost by removing less important prompt tokens. Description-free tool planning can be viewed as an extreme fixed-catalog variant of prompt compression: rather than compressing the tool catalog, we remove the entire serialized schema from the inference prompt. The missing catalog information must therefore be recovered from the fine-tuned model weights rather than from a shorter prompt. Unlike standard knowledge distillation \cite{b12}, we do not require teacher model inference at test time.

\subsection{Parameter-Efficient Fine-Tuning and Retention}

LoRA \cite{b4} freezes the pretrained model and trains low-rank adapter updates, while QLoRA \cite{b3} further reduces memory by quantizing the frozen base model. Although adapter-based fine-tuning generally reduces catastrophic forgetting compared with full fine-tuning \cite{b7}, adapter updates can still bias the effective model toward the fine-tuning distribution and degrade prior capabilities. This is especially relevant for tool planning, where the training distribution consists of structured plans rather than general natural language. We therefore evaluate both tool-planning quality and general multiple-choice retention after fine-tuning.

\subsection{Industrial Tool-Planning Benchmarks}

We use AssetOpsBench \cite{b2} because it provides a bounded but realistic industrial tool-planning environment with domain-specific MCP servers, simulated telemetry, human-authored scenarios, and multi-step workflows. This differs from broad benchmarks such as ToolBench \cite{b10} and MCP-Bench \cite{b19}, which evaluate large-scale tool discovery or cross-domain MCP usage. Our goal is not broad generalization to unseen tools, but controlled evaluation of whether a stable MCP-style catalog can be internalized into a small model.

\section{Problem Definition}
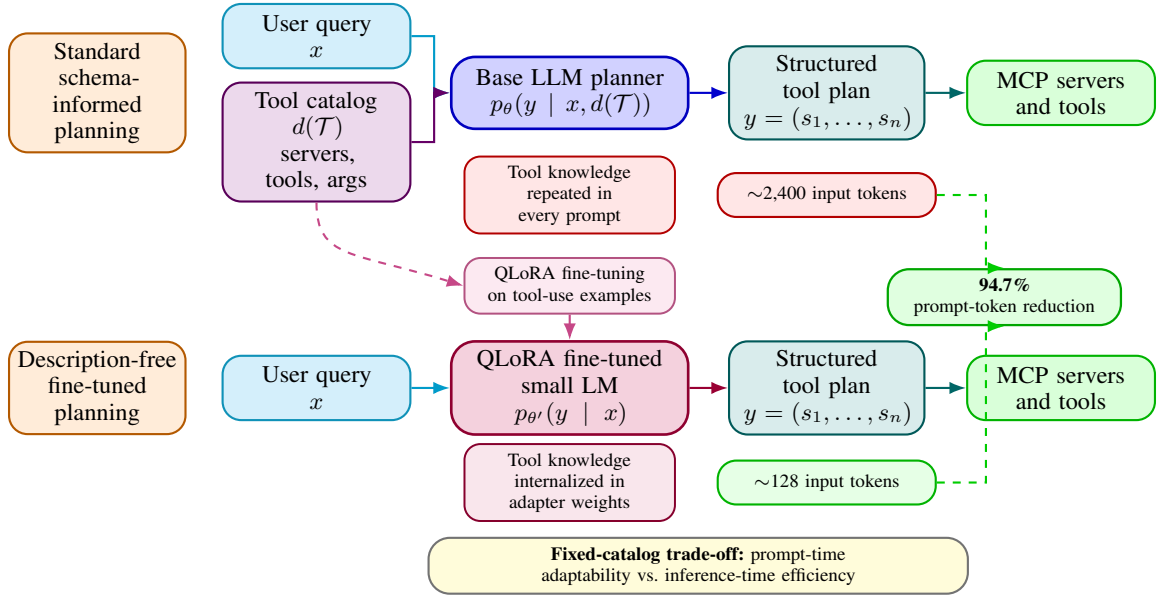
\begin{figure*}[t]
\centering
\begin{tikzpicture}[
    font=\small,
    >=Latex,
    box/.style={
        draw=#1!70!black,
        fill=#1!15,
        rounded corners=7pt,
        align=center,
        minimum height=0.85cm,
        text width=2.25cm,
        line width=0.8pt
    },
    modelbox/.style={
        draw=#1!75!black,
        fill=#1!18,
        rounded corners=9pt,
        align=center,
        minimum height=0.95cm,
        text width=2.9cm,
        line width=1.0pt
    },
    smallbox/.style={
        draw=#1!70!black,
        fill=#1!10,
        rounded corners=6pt,
        align=center,
        font=\scriptsize,
        minimum height=0.58cm,
        text width=2.55cm,
        line width=0.75pt
    },
    callout/.style={
        draw=#1!70!black,
        fill=#1!10,
        rounded corners=7pt,
        align=center,
        font=\scriptsize,
        minimum height=0.58cm,
        text width=2.65cm,
        line width=0.8pt
    },
    arrow/.style={
        -{Latex[length=2.2mm]},
        thick,
        draw=#1!80!black
    },
    dashedarrow/.style={
        -{Latex[length=2.2mm]},
        thick,
        dashed,
        draw=#1!80!black
    }
]

\node[box=orange, text width=2.1cm] (label1) at (0,0)
{Standard\\schema-informed\\planning};

\node[box=orange, text width=2.1cm] (label2) at (0,-3.9)
{Description-free\\fine-tuned\\planning};

\node[box=cyan] (query1) at (2.9,0.75)
{User query\\$x$};

\node[box=violet, minimum height=1.25cm] (catalog1) at (2.9,-0.65)
{Tool catalog\\$d(\mathcal{T})$\\servers, tools, args};

\node[modelbox=blue] (llm1) at (6.25,0)
{Base LLM planner\\$p_{\theta}(y \mid x,d(\mathcal{T}))$};

\node[box=teal, text width=2.35cm] (plan1) at (9.65,0)
{Structured\\tool plan\\$y=(s_1,\ldots,s_n)$};

\node[box=green, text width=2.25cm] (tools1) at (12.75,0)
{MCP servers\\and tools};

\node[smallbox=red] (repeat1) at (6.25,-1.35)
{Tool knowledge\\repeated in\\every prompt};

\node[callout=red] (tokens1) at (9.65,-1.35)
{$\sim$2,400 input tokens};

\draw[arrow=cyan] (query1.east) -- ++(0.3,0) |- (llm1.west);
\draw[arrow=violet] (catalog1.east) -- ++(0.3,0) |- (llm1.west);
\draw[arrow=blue] (llm1.east) -- (plan1.west);
\draw[arrow=teal] (plan1.east) -- (tools1.west);

\node[smallbox=magenta] (train) at (6.25,-2.55)
{QLoRA fine-tuning\\on tool-use examples};

\draw[dashedarrow=magenta]
    (catalog1.south) .. controls +(0,-0.75) and +(-1.45,0.45) .. (train.west);

\node[box=cyan] (query2) at (2.9,-3.9)
{User query\\$x$};

\node[modelbox=purple] (llm2) at (6.25,-3.9)
{QLoRA fine-tuned\\small LM\\$p_{\theta'}(y \mid x)$};

\node[box=teal, text width=2.35cm] (plan2) at (9.65,-3.9)
{Structured\\tool plan\\$y=(s_1,\ldots,s_n)$};

\node[box=green, text width=2.25cm] (tools2) at (12.75,-3.9)
{MCP servers\\and tools};

\node[smallbox=purple] (internal1) at (6.25,-5.15)
{Tool knowledge\\internalized in\\adapter weights};

\node[callout=green] (tokens2) at (9.65,-5.15)
{$\sim$128 input tokens};

\draw[arrow=cyan] (query2.east) -- (llm2.west);
\draw[arrow=magenta] (train.south) -- (llm2.north);
\draw[arrow=purple] (llm2.east) -- (plan2.west);
\draw[arrow=teal] (plan2.east) -- (tools2.west);

\node[
    draw=green!65!black,
    fill=green!12,
    rounded corners=8pt,
    align=center,
    font=\scriptsize,
    text width=2.85cm,
    minimum height=0.7cm,
    line width=0.85pt
] (reduction) at (12.0,-2.7)
{\textbf{94.7\%}\\prompt-token reduction};

\draw[dashedarrow=green] (tokens1.east) -- ++(0.65,0) |- (reduction.north);
\draw[dashedarrow=green] (tokens2.east) -- ++(0.65,0) |- (reduction.south);

\node[
    draw=black!55,
    fill=yellow!18,
    rounded corners=8pt,
    align=center,
    font=\scriptsize,
    minimum height=0.72cm,
    text width=6.9cm,
    line width=0.8pt
] at (7.95,-6.25)
{\textbf{Fixed-catalog trade-off:} prompt-time adaptability vs. inference-time efficiency};

\end{tikzpicture}

\caption{Schema-informed versus description-free tool planning. Standard prompting supplies the user query and serialized tool catalog $d(\mathcal{T})$ at each call, while QLoRA internalizes fixed-catalog tool knowledge into adapter weights, enabling MCP tool-use planning from the query alone.}
\label{fig:framework}
\end{figure*}

We formalize the setting studied in this paper as \emph{description-free tool planning} for a fixed tool catalog. Let $\mathcal{T}$ denote a tool catalog containing a set of MCP servers, tools, argument schemas, return types, and natural-language tool descriptions. A user query is denoted by $x$, and the desired output is a structured plan $y = (s_1, \ldots, s_n)$ consisting of ordered planning steps. Each step $s_i$ specifies a task description, an MCP server, a tool, a JSON argument object, dependencies on previous steps, and an expected output.

In standard schema-informed tool planning, the model receives both the user query and a serialized representation of the tool catalog, denoted $d(\mathcal{T})$. The model therefore generates plans according to

\begin{equation}
p_{\theta}(y \mid x, d(\mathcal{T})),
\end{equation}

where $\theta$ are the parameters of the pretrained model. This is the common prompting paradigm used by many tool-use systems: every query is accompanied by the full or partial tool catalog so the model can infer which tools exist and how they should be called.

In contrast, description-free tool planning removes the serialized catalog from the inference-time prompt. The model must instead generate a valid plan using only the user query and the structured output format:

\begin{equation}
p_{\theta'}(y \mid x),
\end{equation}

where $\theta'$ denotes the parameters of the model after tool-use fine-tuning. The goal is for $\theta'$ to encode enough information about $\mathcal{T}$ to recover correct server routing, tool selection, argument construction, and dependency ordering without seeing $d(\mathcal{T})$ at inference time.

This setting differs from general open-ended tool use in an important way: the tool catalog is assumed to be fixed during evaluation. We do not attempt to generalize to unseen tools. Instead, we study whether knowledge of a stable catalog can be shifted from prompt context into model parameters. This creates a trade-off between adaptability and efficiency. Prompt-based methods can immediately support new tools by adding their descriptions to the prompt, while description-free fine-tuning reduces prompt length and inference overhead but requires additional training or adapter updates when the tool catalog changes.

Following the AssetOpsBench output format, generated plans use the field name \texttt{\#Agent} to identify the responsible component. In our MCP setting, this field corresponds to the MCP server responsible for the tool call. We use ``server'' in prose and preserve \texttt{\#Agent} when referring to the literal output format.

We evaluate this setting using four criteria. First, the generated plan must select the correct MCP servers and tools. Second, it must construct valid argument keys and values for each selected tool. Third, it must order tool calls into a coherent dependency structure. Fourth, it should achieve these objectives while reducing prompt length relative to schema-informed prompting. The central question of this work is therefore:

\begin{quote}
Can a small language model internalize a fixed tool catalog through parameter-efficient fine-tuning, allowing it to outperform schema-informed prompting while omitting tool descriptions at inference time?
\end{quote}

\section{Methodology}

We evaluate whether supervised fine-tuning can substitute for prompt-time tool descriptions in a fixed-catalog setting. The experiments compare schema-informed prompting, description-free prompting without fine-tuning, and description-free prompting after QLoRA fine-tuning. We then analyze adapter rank and capability retention to quantify the cost of specialization.

\subsection{Benchmark and Data}

\paragraph{Catalog-grounded description-free SFT.}
We use a teacher--student data construction protocol. During data construction, the teacher model (Gemini 2.5 Flash \cite{b5}) receives the full serialized catalog $d(\mathcal{T})$ and produces schema facts, question-to-plan mappings, and execution-style traces. During student training and evaluation, the serialized catalog is removed from the prompt. The student observes only the user query, output-format instructions, and target plan, forcing the adapter to encode tool names, server ownership, argument keys, and common dependency patterns in its parameters rather than relying on prompt-time schemas.

\paragraph{AssetOpsBench.} We use AssetOpsBench \cite{b2} as the primary benchmark for both data construction and evaluation. The benchmark contains 152 natural-language scenarios, each requiring the model to select appropriate domain tools, generate valid arguments, and order tool calls into a structured execution plan. Each plan consists of sequenced steps in a structured format specifying a task description, the assigned MCP server, the tool to call, JSON arguments, and inter-step dependencies. Table~\ref{tab:compact-tool-inventory} gives a compact overview of the MCP servers and tool counts.

\begin{table}[t]
\small
\centering
\caption{AssetOpsBench MCP server inventory used in this work}
\label{tab:compact-tool-inventory}
\begin{tabular}{lr}
\toprule
\textbf{MCP Server} & \textbf{Tools} \\
\midrule
IoT & 4 \\
FMSR & 2 \\
TSFM & 6 \\
Utilities & 3 \\
WorkOrder & 8 \\
\bottomrule
\end{tabular}
\end{table}

\paragraph{Training data.} The resulting data comprises three types of supervised fine-tuning examples (see Appendix~\ref{app:data} for composition). \emph{Tool and server knowledge} examples teach the model which MCP servers and tools exist, how they are associated, what arguments each tool requires, and how to distinguish between near-miss tools such as \texttt{get\_failure\_modes} and \texttt{get\_failure\_mode\_sensor\_mapping}. \emph{Question-to-plan} examples, based on the 152 AssetOpsBench scenarios together with generated paraphrases, teach the model to map a natural-language query to a structured plan; each plan is validated to follow the required \texttt{\#Task}/\texttt{\#Agent}/\texttt{\#Tool}/\texttt{\#Args} format. \emph{Execution-style} examples combine planning steps with execution traces and placeholder-resolution patterns such as \texttt{\{step\_N\}}.

\paragraph{Prompt structure.} At inference time, each query is embedded in a structured planning prompt (shown in full in Appendix~\ref{app:prompt-example}). The prompt contains four components: a system preamble describing the model's role as a planning assistant; the complete tool catalog listing all MCP server names, tool signatures, argument types, and descriptions ($\sim$2,200 tokens); the structured output format specification
(\texttt{\#Task}/%
\allowbreak\texttt{\#Agent}/%
\allowbreak\texttt{\#Tool}/%
\allowbreak\texttt{\#Args}/%
\allowbreak\texttt{\#Dependency}/%
\allowbreak\texttt{\#ExpectedOutput});
and the user question. The full prompt totals approximately 2,400 tokens. In the description-free setting, we remove the complete tool-catalog component while keeping the system preamble, output-format specification, and user question, reducing the prompt to $\sim$128 tokens---a 94.7\% prompt-token reduction.

\paragraph{Data split.} We use a pattern-aware stratified 80/20 split over the 152 scenarios, where each scenario is assigned a pattern based on its dominant task family, tool set, and step count. The split is performed at the scenario level \emph{before} paraphrase expansion, so all paraphrases and derived traces for a given scenario remain exclusively in either training or test. This produces 122 training scenarios and 30 held-out test scenarios. The full training corpus contains approximately 1,833 examples. We define three training configurations: Config~A (Plan-only, $\sim$1,200 examples), Config~B (Tool-knowledge only, $\sim$500 examples), and Config~C (Tool+Plan, $\sim$1,741 examples). We use Config~C as the primary configuration because it provides the broadest tool-knowledge coverage. Data-composition ablations are in Appendix~\ref{app:data-ablation}.

\subsection{Models and Training}

We compare two open-source instruction-tuned models: Gemma 4 E4B-it \cite{b6} and Qwen3-4B \cite{b17}. Table~\ref{tab:model-summary} summarizes their architectural differences. Both share a hidden size of 2560 but differ in depth, attention mechanism, and vocabulary size.

\begin{table}[t]
\small
\centering
\caption{Base models used for fine-tuning}
\label{tab:model-summary}
\begin{tabular}{lcc}
\toprule
 & \textbf{Gemma 4 E4B} & \textbf{Qwen3-4B} \\
\midrule
Total params & $\sim$8B w/ PLE & 4.0B \\
Active params & 4.5B & 3.6B \\
Layers & 42 & 36 \\
Attention & Hybrid & GQA \\
Hidden size & 2560 & 2560 \\
Vocab size & 262K & 152K \\
\bottomrule
\end{tabular}
\end{table}

Both models are fine-tuned using QLoRA with 8-bit quantization, which keeps the base model weights fixed while training a small set of adapter parameters. We use the same Config~C dataset for both models (1,741 training, 92 evaluation examples) and train with identical hyperparameters unless otherwise stated: LoRA rank $r=32$, $\alpha=64$, dropout $0.05$, learning rate $2 \times 10^{-4}$ with cosine schedule, batch size $2$ with gradient accumulation of $4$, $2$ epochs, early stopping with patience $2$, and $436$ total training steps. All linear layers are targeted.

Although the two models use the same LoRA rank, the resulting trainable-parameter fraction differs due to model-size differences. For Gemma, rank $32$ corresponds to approximately $0.63\%$ trainable parameters, whereas for Qwen3 it corresponds to approximately $1.62\%$ and about $2.6\times$ higher adaptation intensity. This difference directly affects retention, as discussed in Section~\ref{sec:retention}.

\subsection{Evaluation Protocol}
\label{sec:evaluation}

We evaluate plans using two complementary approaches. Structural metrics provide deterministic, reproducible scores but can penalize valid plans that deviate from the gold reference. LLM-based judging captures semantic quality and accepts valid alternative formulations, but may exhibit grading bias. Using both approaches and checking for agreement provides more robust evidence than either alone.

For structural evaluation, we report AT-F1, which extracts the set of (MCP server, tool) pairs from actionable steps (excluding steps where agent or tool is ``none'') in both the gold and candidate plans and computes set-based F1 (precision $=$ matched pairs / candidate pairs, recall $=$ matched pairs / gold pairs). We also report ArgKey-F1, which measures whether the model predicts the correct argument field names for matched tool calls. For semantic evaluation, we use Gemini 2.5 Flash as an LLM-as-judge \cite{b13} to rate each plan on six dimensions on a 1--5 scale: correctness, server routing, tool selection, argument quality, efficiency, and dependency correctness. The judge overall score is the mean of all six dimensions. Full judge prompts are in Appendix~\ref{app:judge}.

We evaluate models under three prompting conditions: \emph{informed}, where full tool descriptions are included in the prompt; \emph{description-free}, where the base model receives only the user query with no tool schemas; and \emph{fine-tuned description-free}, where the fine-tuned model generates plans without tool schemas at inference time.

We also evaluate capability retention using 100 multiple-choice questions drawn from MMLU \cite{b14}, ARC-Challenge \cite{b15}, and HellaSwag \cite{b16}. This measures whether tool-use specialization degrades general reasoning behavior that may still be needed by a multi-purpose planner. Details are in Appendix~\ref{app:retention}.

\section{Experimental Results}

\subsection{Main Result: Description-Free Fine-Tuning Improves Tool Planning}
\label{sec:main-result}

We first evaluate whether fine-tuning can replace prompt-time tool descriptions. The unfine-tuned Gemma~4~E4B model is evaluated in two conditions. In the \emph{informed} condition, the prompt contains the system preamble, full serialized tool catalog ($\sim$2,200 tokens of schema text), the output format specification, and the user query, totaling approximately 2,400 input tokens per query (see Appendix~\ref{app:prompt-example}). In the \emph{description-free} condition, the same unfine-tuned model receives only the system preamble, output format, and user query ($\sim$128 tokens). The description-free baseline achieves zero AT-F1, confirming that the base model cannot reliably produce valid tool-use plans without explicit tool descriptions.

\begin{figure}[t]
\centering
\includegraphics[width=\columnwidth]{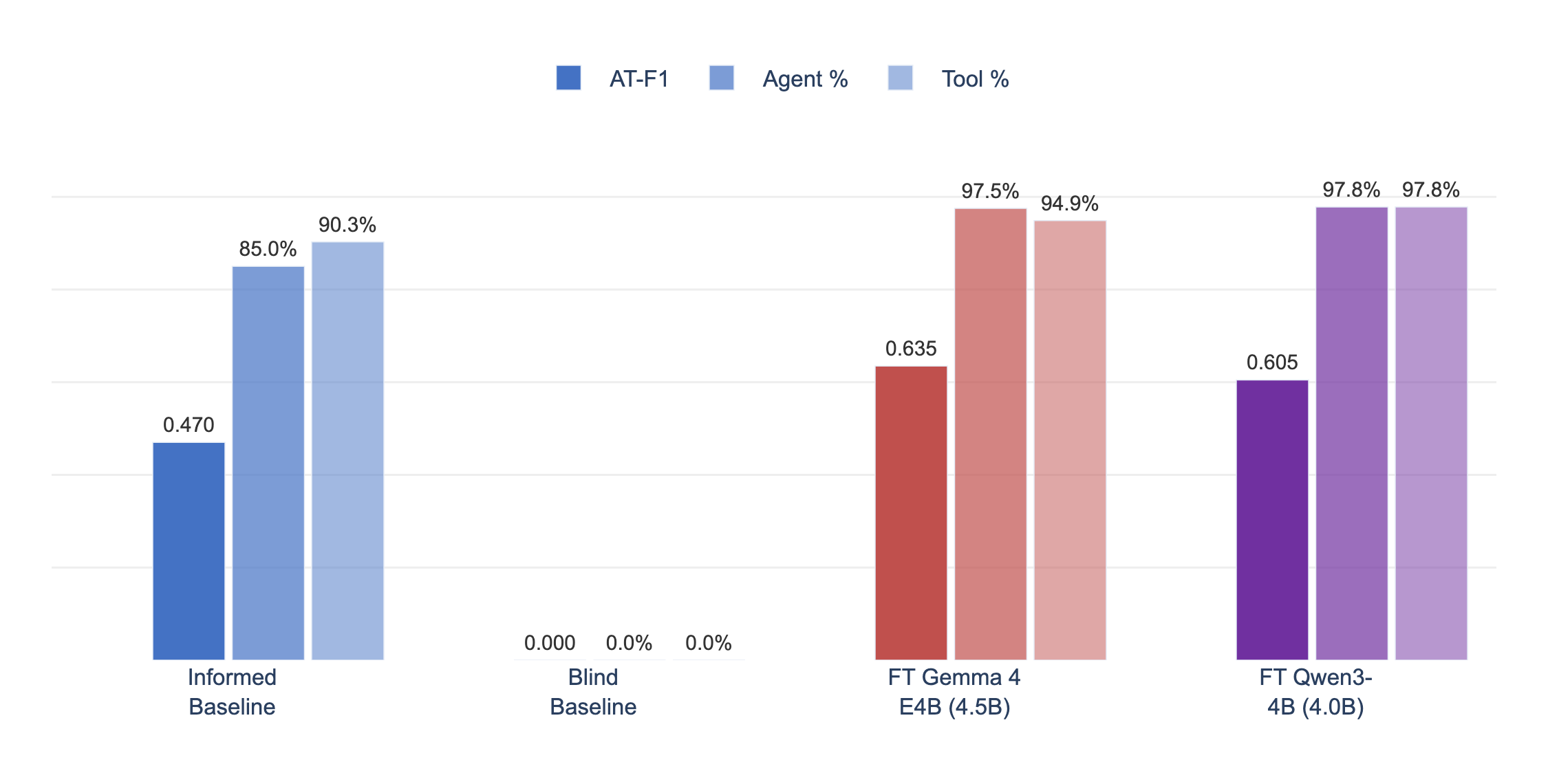}
\caption{Structural planning metrics across all four configurations. Fine-tuned models operating without tool descriptions surpass the informed baseline on AT-F1, server routing, and tool selection.}
\label{fig:main-results}
\end{figure}

\begin{figure}[t]
\centering
\includegraphics[width=\columnwidth]{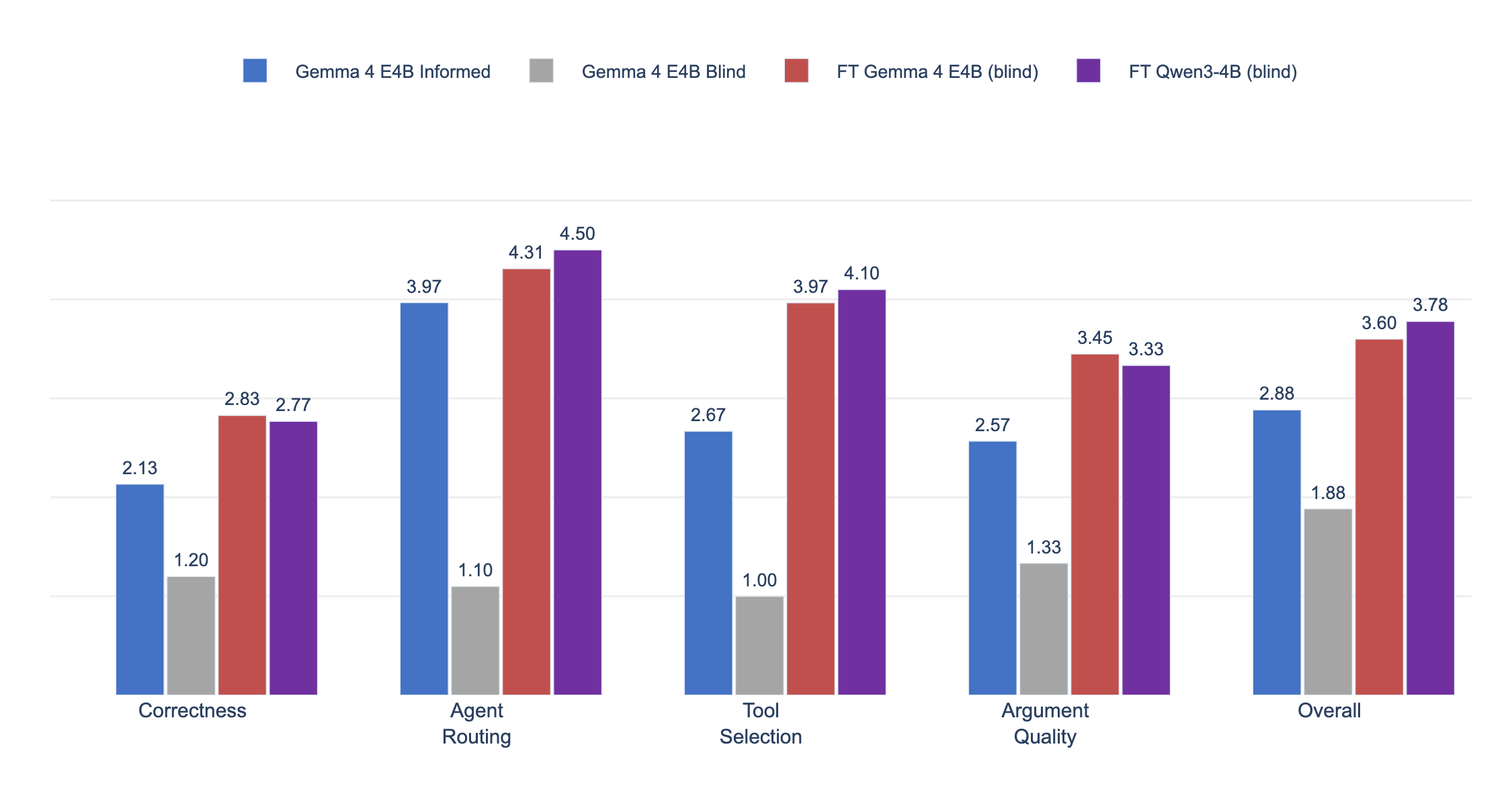}
\caption{LLM-as-judge scores (1--5 scale) across five evaluation dimensions. Fine-tuned models operating without tool descriptions outperform the informed baseline on every dimension.}
\label{fig:judge-scores}
\end{figure}

Fine-tuning changes this substantially. Figs.~\ref{fig:main-results} and~\ref{fig:judge-scores} show that fine-tuned models perform well in the description-free setting and outperform the informed baseline on both structural and semantic evaluation metrics.

The fine-tuned Gemma model (Config~C, $r$=32) reaches an AT-F1 of 0.635 and an overall judge score of 3.60, compared with 0.47 and 2.88 for the informed baseline. Server routing and tool-selection accuracy remain high, reaching approximately 95--98\% across the held-out evaluation set. The fine-tuned Qwen3 model achieves an AT-F1 of 0.605 and an overall judge score of 3.78, while using substantially less memory. Both models substantially outperform the informed baseline, confirming that fine-tuned description-free planning is viable. As shown in the rank sweep (Section~\ref{sec:lora-rank}), a separate Gemma training run with the same configuration achieves a judge score of 3.88, indicating that the improvement over the informed baseline is robust across runs despite some run-to-run variance on the 30-scenario test set.

\textbf{Metric agreement.} Structural and judge-based metrics agree directionally: both rank the configurations in the same broad order---description-free baseline worst, informed baseline substantially better, and fine-tuned description-free models best. This agreement is useful because AT-F1 is deterministic but strict, while LLM judging captures semantically valid alternatives but may introduce grading bias.

Because the 1--5 judge scale has a floor of 1.0, the description-free baseline's score of 1.88 corresponds to 0.88/4.0 on a zero-anchored scale, confirming near-zero semantic planning quality in the absence of either tool descriptions or fine-tuning.

\begin{figure}[t]
\centering
\includegraphics[width=\columnwidth]{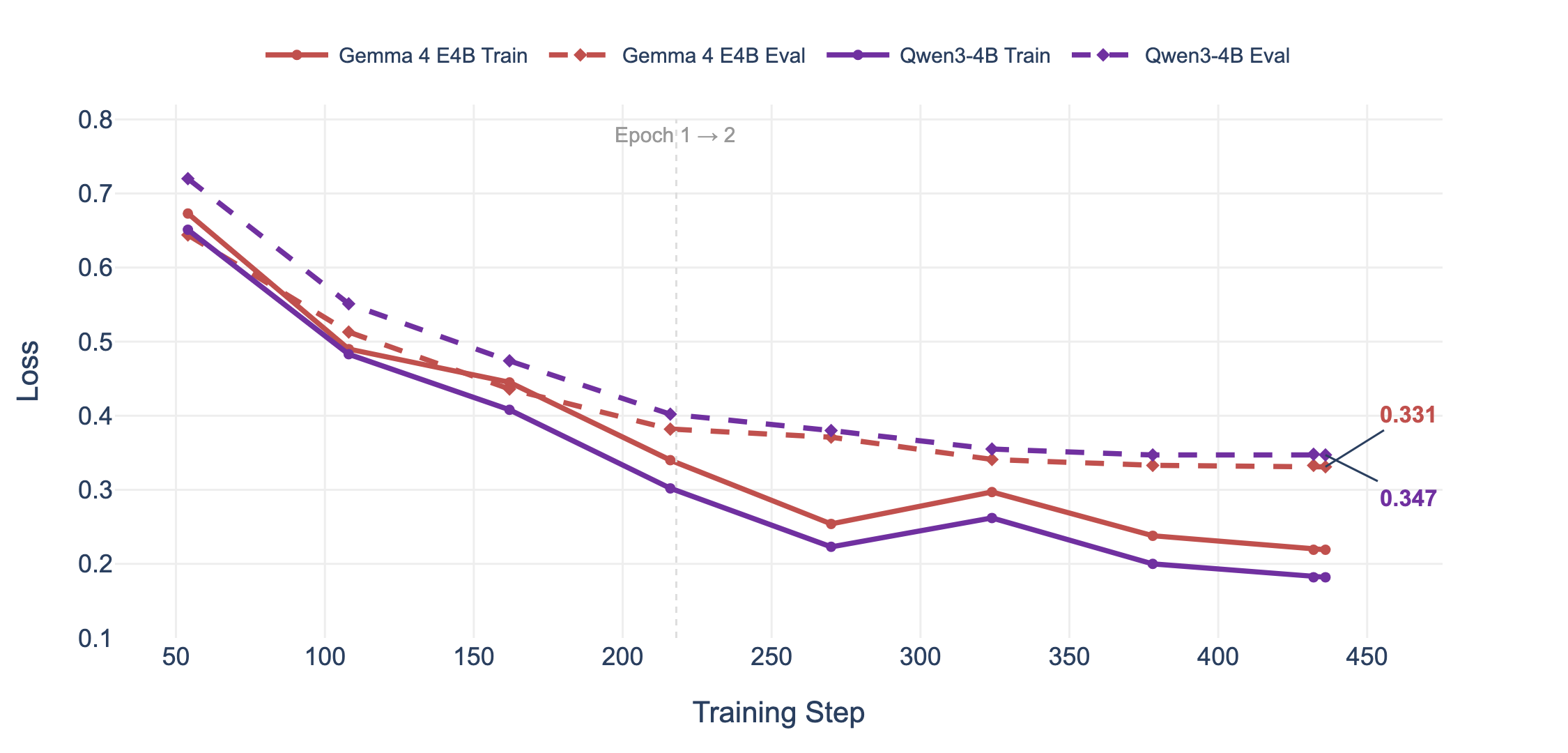}
\caption{Training and evaluation loss over 436 steps (2 epochs). Both models converge smoothly to comparable final evaluation losses (Gemma 0.331, Qwen3 0.347), despite Qwen3 starting from a higher initial loss.}
\label{fig:training-curves}
\end{figure}

Both models converge smoothly within two epochs (Fig.~\ref{fig:training-curves}), reaching comparable evaluation losses (Gemma 0.331; Qwen3 0.347). The prompt-token savings are substantial: description-free inference after fine-tuning reduces the planning prompt from $\sim$2,400 tokens to $\sim$128 tokens, a 94.7\% reduction.

\subsection{LoRA Rank Selection}
\label{sec:lora-rank}

Having established that fine-tuning enables description-free inference that surpasses the informed baseline, we next study how adapter capacity affects planning quality. LoRA rank $r$ controls the number of trainable parameters and thus the model's capacity to absorb new tool-use knowledge. We sweep $r \in \{8, 16, 32, 64\}$ on Gemma~4~E4B, corresponding to trainable-parameter fractions from 0.32\% to 2.51\%.

\begin{figure}[t]
\centering
\includegraphics[width=\columnwidth]{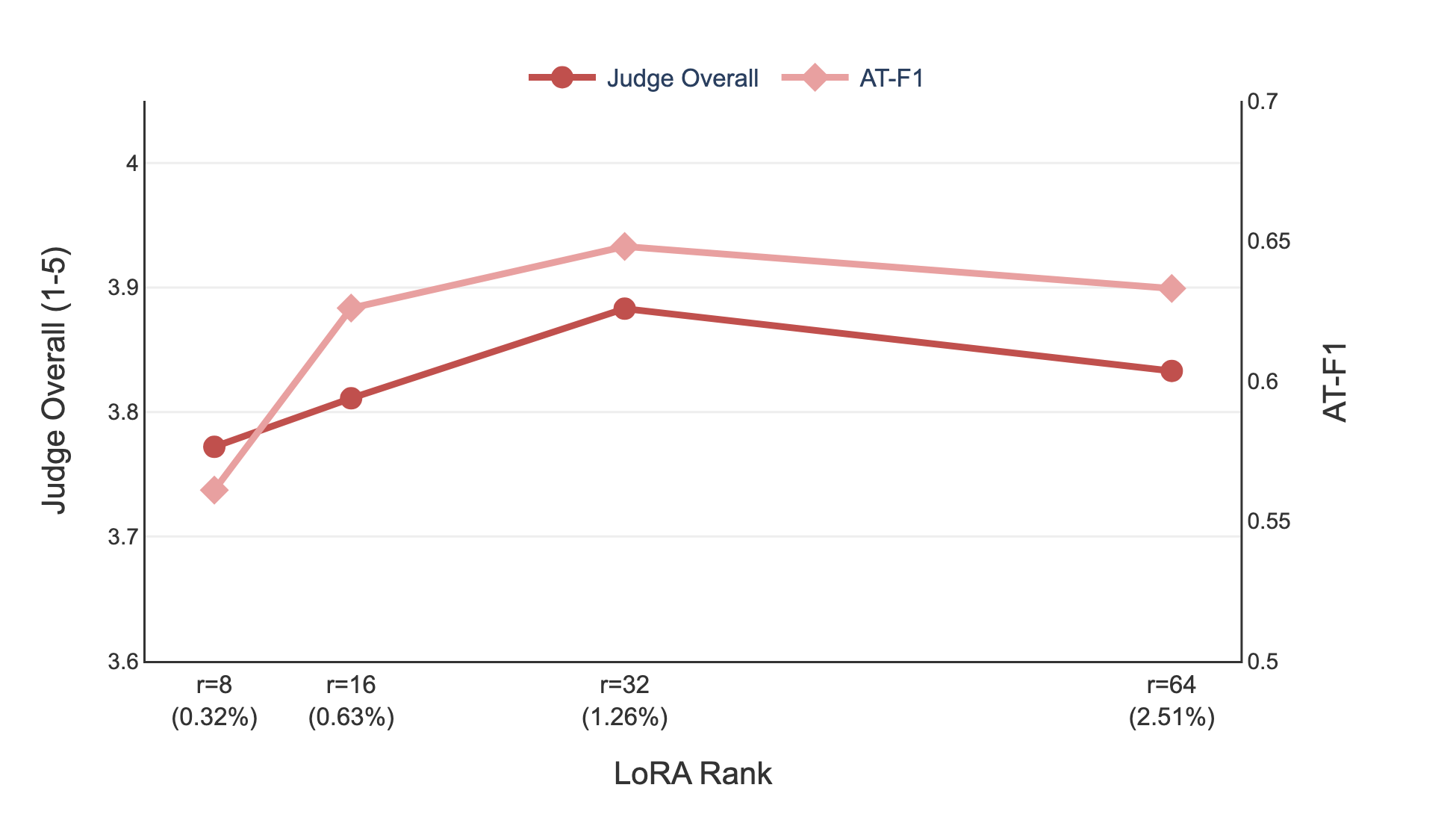}
\caption{Effect of LoRA rank on planning quality (Gemma~4~E4B). Both Judge Overall (left axis) and AT-F1 (right axis) peak at $r$=32 and decline at $r$=64, suggesting diminishing returns at higher adapter capacity.}
\label{fig:lora-rank}
\end{figure}

As shown in Fig.~\ref{fig:lora-rank}, both metrics rise steeply from $r$=8 (Judge 3.77, AT-F1 0.56) to $r$=16 (Judge 3.81, AT-F1 0.63) and peak at $r$=32 (Judge 3.88, AT-F1 0.65). At $r$=64, performance slightly degrades (Judge 3.83, AT-F1 0.63), suggesting diminishing returns at higher adapter capacity. This finding is important for the retention analysis in Section~\ref{sec:retention}: while $r$=32 maximizes planning quality, smaller ranks preserve more general knowledge.

\subsection{Capability Retention}
\label{sec:retention}

A planner deployed inside a broader agent may still need to recognize unrelated requests, answer simple questions, or reason outside the tool catalog. We therefore measure whether tool-use fine-tuning degrades general multiple-choice performance. We use 100 questions from MMLU, ARC-Challenge, and HellaSwag, independent of AssetOpsBench. We compare Gemma at $r$=8 and $r$=32 to test the rank trade-off, and Qwen3 at $r$=32.

\begin{figure}[t]
\centering
\includegraphics[width=\columnwidth]{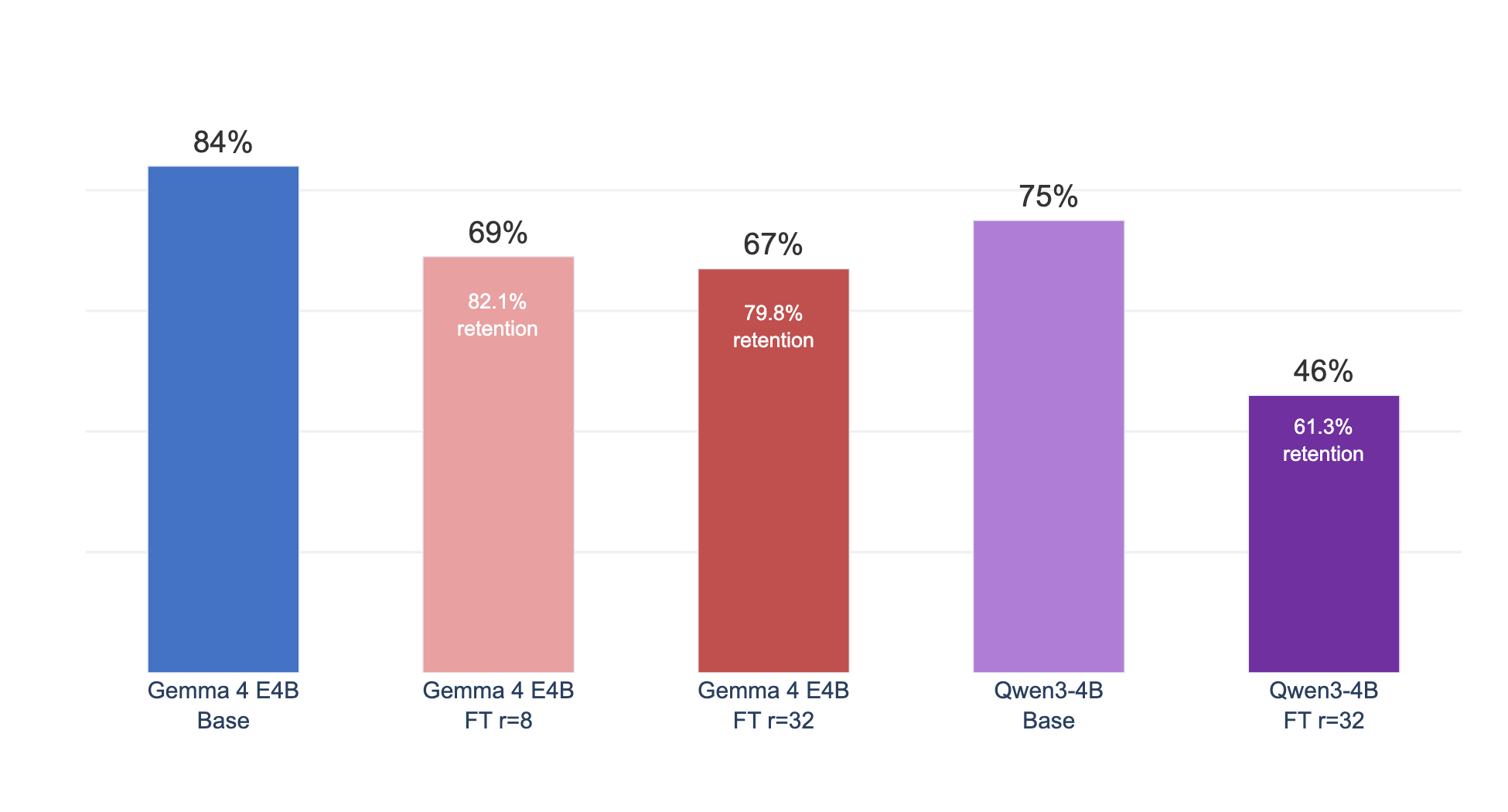}
\caption{Overall MCQ accuracy before and after tool-use fine-tuning. Gemma~4~E4B retains 79.8--82.1\% of base performance; Qwen3-4B retains only 61.3\%. Retention percentages shown inside bars. Per-benchmark breakdown in Appendix~\ref{app:retention}.}
\label{fig:retention}
\end{figure}

As shown in Fig.~\ref{fig:retention}, Gemma shows moderate degradation: the base model scores 84.0\% MCQ accuracy, dropping to 69.0\% at $r$=8 (82.1\% retention) and 67.0\% at $r$=32 (79.8\% retention). Per-question analysis reveals that Gemma at $r$=32 forgot 21 of 100 questions while learning 4 new ones. This reveals a quality--retention trade-off: Section~\ref{sec:lora-rank} showed that $r$=32 achieves the best judge score (3.88 vs.\ 3.77 for $r$=8), but at the cost of 2.3 percentage points of retention. For deployment where general reasoning matters, the smaller $r$=8 adapter may be preferable.

Qwen3 shows larger degradation: its base accuracy drops from 75.0\% to 46.0\% after fine-tuning at $r$=32, or 61.3\% retention. One likely contributor is adaptation intensity: at the same LoRA rank, Qwen3 trains 1.62\% of its parameters compared with 0.63\% for Gemma. However, architecture, tokenizer, and pretraining differences are confounded, so this should not be interpreted as a causal explanation. The per-benchmark breakdown (Fig.~\ref{fig:retention-detail} in Appendix~\ref{app:retention}) reveals that Qwen3's degradation is especially severe on logical reasoning (75\%$\rightarrow$25\%) and commonsense completion (50\%$\rightarrow$27\%), while Gemma's degradation is more evenly distributed.

We verified that degradation was not caused by inference quantization: the base Gemma model achieved 84\% in bf16 and 87\% in 8-bit, confirming that quantization alone does not explain the 84\%$\rightarrow$67\% drop observed after fine-tuning.

\subsection{Deployment Characterization}
\label{sec:profiling}

We report basic deployment measurements to characterize the practical cost of description-free fine-tuning. Table~\ref{tab:profiling-comparison} summarizes the results on a single A100 80GB GPU.

\begin{table}[t]
\small
\centering
\caption{Profiling comparison on A100 80GB}
\label{tab:profiling-comparison}
\begin{tabular}{lcc}
\toprule
\textbf{Metric} & \textbf{Gemma} & \textbf{Qwen3} \\
\midrule
Base memory & 11.5\,GB & 4.42\,GB \\
Peak train memory & 24.1\,GB & 16.06\,GB \\
Train time & 56\,min & 39.7\,min \\
Inference speed & 1.4\,tok/s & 3.49\,tok/s \\
Eval time & 59\,min & 40.7\,min \\
Best eval loss & 0.331 & 0.347 \\
\bottomrule
\end{tabular}
\end{table}

Qwen3 uses 62\% less base memory, trains 29\% faster, and runs 2.5$\times$ faster at inference while achieving a comparable planning judge score. Qwen3's speed advantage comes directly from having fewer total parameters (4.0B vs.\ 8.0B), which reduces the size of every 8-bit matrix multiplication. Full profiling details, including the CUDA operator breakdown and cost estimation, are provided in Appendix~\ref{app:profiling}.

Additional data-composition and quantization ablations are provided in Appendices~\ref{app:data-ablation} and~\ref{app:quant}.

\section{Discussion}

\paragraph{Fixed-catalog internalization.}
The main result shows that small models can learn a fixed tool catalog well enough to plan without prompt-time tool descriptions. This does not replace schema-informed prompting for open-world or rapidly changing tools. Rather, it targets mature deployments where the same catalog is reused across many queries and schema knowledge can be amortized into adapter weights.

\paragraph{Why can description-free fine-tuning beat schema-informed prompting?}
One surprising result is that fine-tuned description-free models outperform an unfine-tuned model that receives the full schema. We hypothesize two mechanisms. First, long schema prompts create context competition: the serialized tool catalog dominates the input, and small models may fail to attend to the relevant schema fragment. Second, schema parsing is itself an inference-time task. Supervised fine-tuning converts this into a learned mapping from user intents to tool sequences, arguments, and dependencies. We do not directly isolate these mechanisms, and future work should evaluate fine-tuned models both with and without schemas at inference time to separate internalization effects from context-length effects.

\paragraph{Specialization trade-offs.}
The retention results show that tool internalization has a cost. Higher LoRA rank improves planning quality but can degrade general multiple-choice performance. This matters when the planner is part of a multi-purpose agent that must handle both tool-related and unrelated requests. Lower-rank adapters, replay data, or modular adapter architectures may reduce this trade-off. A preliminary 10-scenario end-to-end execution pilot validated that plan quality correlates with task completion (70\% success), but full closed-loop evaluation remains for future work. Future work should also study continual learning for adding new tools without forgetting old ones, larger benchmarks, multi-seed runs with confidence intervals, and optimized quantized kernels such as GPTQ or AWQ.

\section{Conclusion}

We studied description-free tool planning for fixed catalogs, where a model must generate structured MCP tool-use plans without receiving tool descriptions at inference time. On AssetOpsBench, 8-bit QLoRA fine-tuning enables $\sim$4B parameter models to outperform an unfine-tuned schema-informed baseline while reducing prompt length by 94.7\%. The best Gemma run achieves 0.65 AT-F1 and a 3.88 judge score, compared with 0.47 and 2.88 for the informed baseline.

The gains come with a specialization cost: LoRA rank affects both planning quality and capability retention. Gemma retains approximately 80--82\% of its base MCQ performance after fine-tuning, while Qwen3 shows larger degradation. Overall, these results suggest that supervised adapter fine-tuning is a practical mechanism for amortizing fixed tool-catalog knowledge in small models, but that retention and catalog-update costs remain important deployment considerations.

\section{Limitations}

This study is a proof of concept on AssetOpsBench. The held-out evaluation set contains only 30 scenarios, and most results are based on single fine-tuning runs rather than multiple random seeds. Small differences between model variants should therefore be interpreted cautiously.

Our evaluation focuses on plan quality rather than full closed-loop execution. A deployed agent may fail at execution time even when the generated plan is structurally correct. We do not evaluate generalization to unseen tools; this is by design, as description-free planning assumes a fixed catalog. Benchmark coverage is incomplete: not all AssetOpsBench MCP servers, scenarios, or asset classes are evaluated.

\section{Use of Generative AI}

 We used generative AI tools for limited drafting, editing, proofreading assistance, and exploratory scripting. The authors reviewed, edited, and verified the final manuscript, experiments, results, and references, and take full responsibility for the content of the paper.


\bibliographystyle{IEEEtran}
\bibliography{custom}

\appendix

\section{Training Data Composition}
\label{app:data}

\begin{table}[!htbp]
\centering
\caption{Three training datasets}
\label{tab:three-training-datasets}
\footnotesize
\setlength{\tabcolsep}{3pt}
\renewcommand{\arraystretch}{1.15}
\begin{tabular}{p{0.23\columnwidth}p{0.13\columnwidth}p{0.27\columnwidth}p{0.27\columnwidth}}
\toprule
\textbf{Dataset} & \textbf{N} & \textbf{Source} & \textbf{Description} \\
\midrule
Tool Knowledge
& $\sim$500
& Gemini 2.5 Flash
& Tool taxonomy, ownership, args, routing, hard negatives \\
Planning
& $\sim$1,200
& Gold plans + paraphrases
& Scenario $\rightarrow$ concise planning steps \\
Execution
& $\sim$100
& Gold plans + traces
& Scenario $\rightarrow$ planning + execution links \\
\bottomrule
\end{tabular}

\vspace{0.3em}
\begin{minipage}{0.95\columnwidth}
\footnotesize
\textit{Note.} Total $\sim$1,833 examples. Config~C (Tool+Plan) uses $\sim$1,741 after 95/5 train/eval split.
\end{minipage}
\end{table}

\section{Training Data Ablation}
\label{app:data-ablation}

\begin{figure}[!htbp]
\centering
\includegraphics[width=\columnwidth]{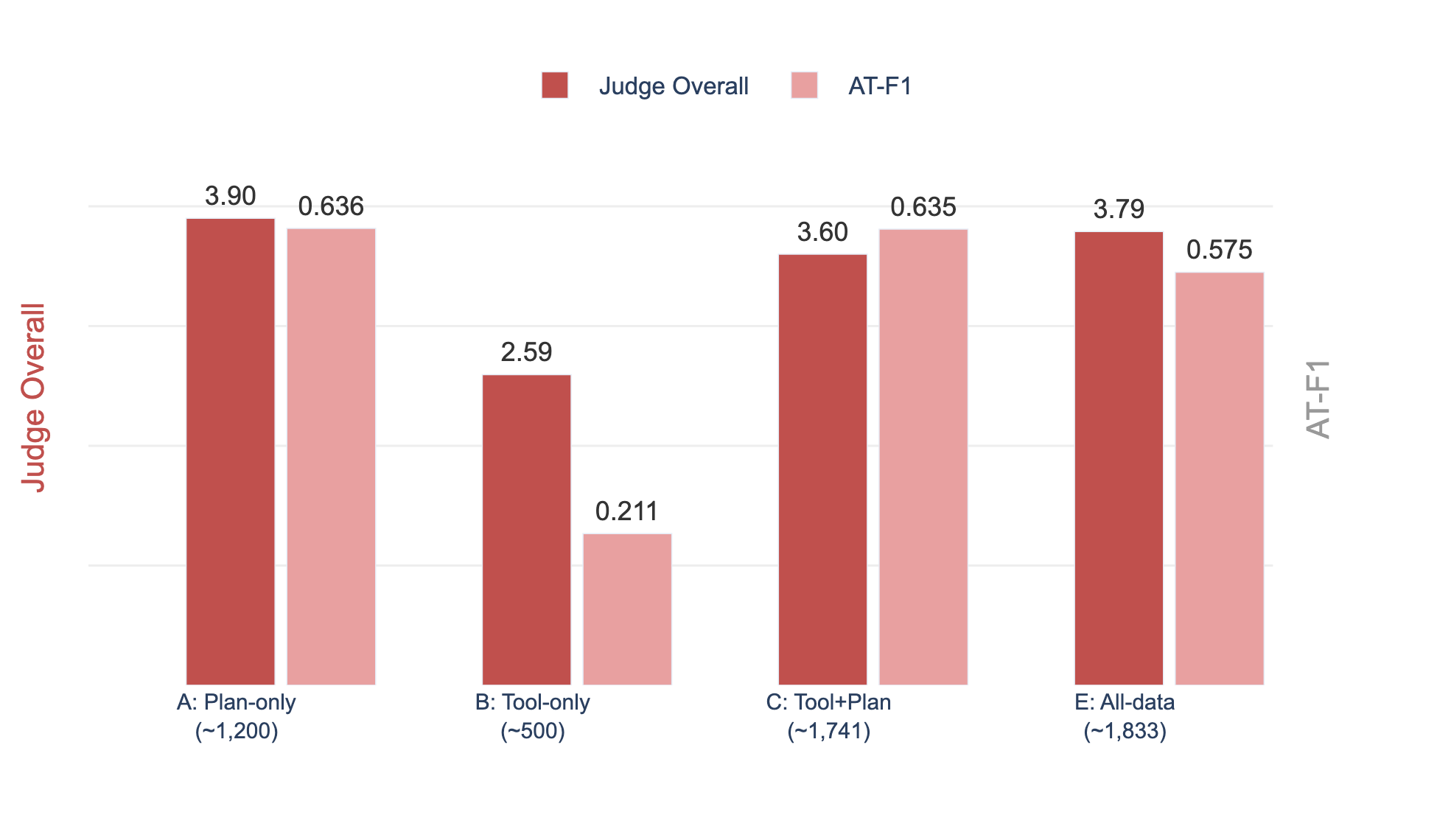}
\caption{Effect of training data composition on planning quality. Plan-only data achieves the highest judge score (3.90), while Tool-only data performs poorly (2.59).}
\label{fig:data-ablation}
\end{figure}

We compare four data configurations using 8-bit QLoRA with $r$=32 on Gemma~4~E4B. Fig.~\ref{fig:data-ablation} shows the results.

Plan-only training (Config~A, $\sim$1,200 examples) achieves the highest judge score of 3.90 and AT-F1 of 0.636. Adding tool-knowledge examples (Config~C) does not improve the judge score in this run (3.60), though tool selection accuracy is higher (94.9\% vs.\ 92.9\%). Tool-only training (Config~B) performs poorly (2.59). One possible explanation is that planning examples expose tool usage in realistic query contexts, whereas tool-knowledge examples are more declarative. We treat these results as exploratory because the configurations also differ in size and coverage, and run-to-run variance on the 30-scenario test set can be substantial (see Limitations).

\FloatBarrier

\section{Quantization Ablation}
\label{app:quant}

\begin{table}[!htbp]
\centering
\caption{Quantization ablation (Gemma 4 E4B, $r$=32)}
\label{tab:quantization-ablation}
\begin{tabular}{lccc}
\toprule
\textbf{Quantization} & \textbf{AT-F1} & \textbf{Judge} & \textbf{Train Time} \\
\midrule
8-bit & 0.617 & 3.78 & 56\,min \\
4-bit NF4 & 0.642 & 3.74 & $\sim$50\,min \\
\bottomrule
\end{tabular}

\vspace{0.3em}
\begin{minipage}{0.95\columnwidth}
\centering
\textit{Note.} Differences are marginal. We use 8-bit as the default for its higher judge score.
\end{minipage}
\end{table}

\section{Profiling Details}
\label{app:profiling}

\begin{figure}[!htbp]
\centering
\includegraphics[width=\columnwidth]{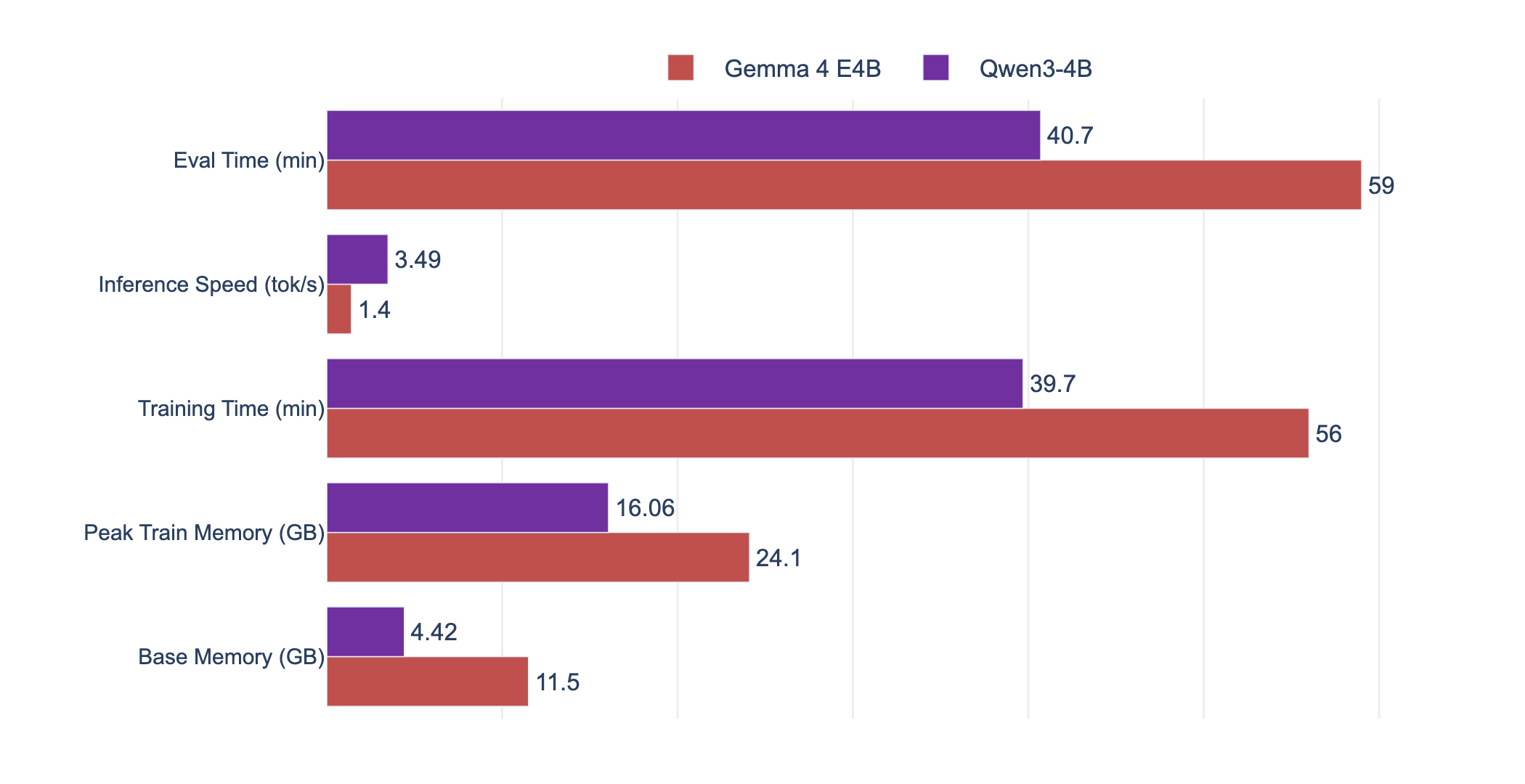}
\caption{Profiling comparison between Gemma~4~E4B and Qwen3-4B. Qwen3 uses less memory, trains faster, and achieves 2.5$\times$ higher inference throughput.}
\label{fig:profiling}
\end{figure}

All profiling was conducted on a single NVIDIA A100 80GB GPU. Fig.~\ref{fig:profiling} provides a visual comparison; Tables~\ref{tab:cuda-breakdown} and~\ref{tab:cost-breakdown} provide additional detail.

For profiling, we use the PyTorch Profiler, \texttt{torch.cuda} memory tracking, and Weights \& Biases. Training throughput scaled from $\sim$886 tok/s at batch size 1 to $\sim$1,415 tok/s at batch size 4, beyond which the model exceeded available memory. The main CUDA bottleneck was \texttt{MatMul8bitLt}, which accounted for 56.3\% of total CUDA time, reflecting the dominance of 8-bit matrix multiplication in quantized inference.

\begin{table}[!htbp]
\centering
\caption{CUDA operator breakdown (Gemma inference, 64 tokens)}
\label{tab:cuda-breakdown}
\begin{tabular}{lc}
\toprule
\textbf{Operation} & \textbf{\% CUDA Time} \\
\midrule
\texttt{MatMul8bitLt} & 56.3\% \\
\texttt{aten::mm} & 18.2\% \\
Attention kernel & 8.7\% \\
Other ops & 16.8\% \\
\bottomrule
\end{tabular}

\vspace{0.3em}
\begin{minipage}{0.95\columnwidth}
\textit{Note.} Attention kernel refers to \texttt{scaled\_dot\_product\_attention}. Other ops include softmax, layer normalization, and related kernels.
\end{minipage}
\end{table}

\begin{table}[!htbp]
\centering
\caption{Experiment cost breakdown (A100 at \$3.93/hr)}
\label{tab:cost-breakdown}
\begin{tabular}{lcc}
\toprule
\textbf{Component} & \textbf{A100 Hours} & \textbf{Cost} \\
\midrule
Gemma ablations & 3.1 & \$12.18 \\
Qwen3 train + eval & 1.5 & \$5.78 \\
Retention analysis & 11.2 & \$44.00 \\
\midrule
\textbf{Total} & \textbf{$\sim$15.8} & \textbf{$\sim$\$62} \\
\bottomrule
\end{tabular}
\end{table}

\FloatBarrier

\section{Judge Prompts}
\label{app:judge}

\subsection{Plan Quality Judge}

We use Gemini 2.5 Flash (temperature=0, max\_tokens=8192) to evaluate each candidate plan against the gold reference. The prompt provides the question, gold plan, candidate plan, and tool inventory, and asks the judge to rate on a 1--5 scale across six dimensions: correctness, server routing, tool selection, argument quality, efficiency, and dependency correctness. Each dimension includes a rubric (5 = matches gold reference, 3 = partially correct, 1 = major errors). The judge returns a JSON object with integer scores.

\subsection{MCQ Retention Judge}

For the retention benchmark, we use Gemini 2.5 Flash (temperature=0, max\_tokens=1024). The prompt provides the question, answer choices, correct answer, and the model's full response. The judge is instructed to look at the \emph{final} answer the model commits to, ignoring intermediate reasoning, and return a JSON object with a binary correct/incorrect grade. This approach is more reliable than token-level evaluation because instruction-tuned models produce chain-of-thought reasoning before committing to an answer letter.

\FloatBarrier

\section{Retention Benchmark Details}
\label{app:retention}

\begin{table}[!htbp]
\centering
\caption{Retention benchmark composition}
\label{tab:retention-benchmark-composition}
\footnotesize
\begin{tabular}{lcl}
\toprule
\textbf{Source} & \textbf{N} & \textbf{Description} \\
\midrule
MMLU (5 subjects) & 40 & Knowledge and reasoning \\
ARC-Challenge & 30 & Grade-school science \\
HellaSwag & 30 & Commonsense completion \\
\midrule
Total & 100 & \\
\bottomrule
\end{tabular}
\end{table}

\begin{figure}[!htbp]
\centering
\includegraphics[width=\columnwidth]{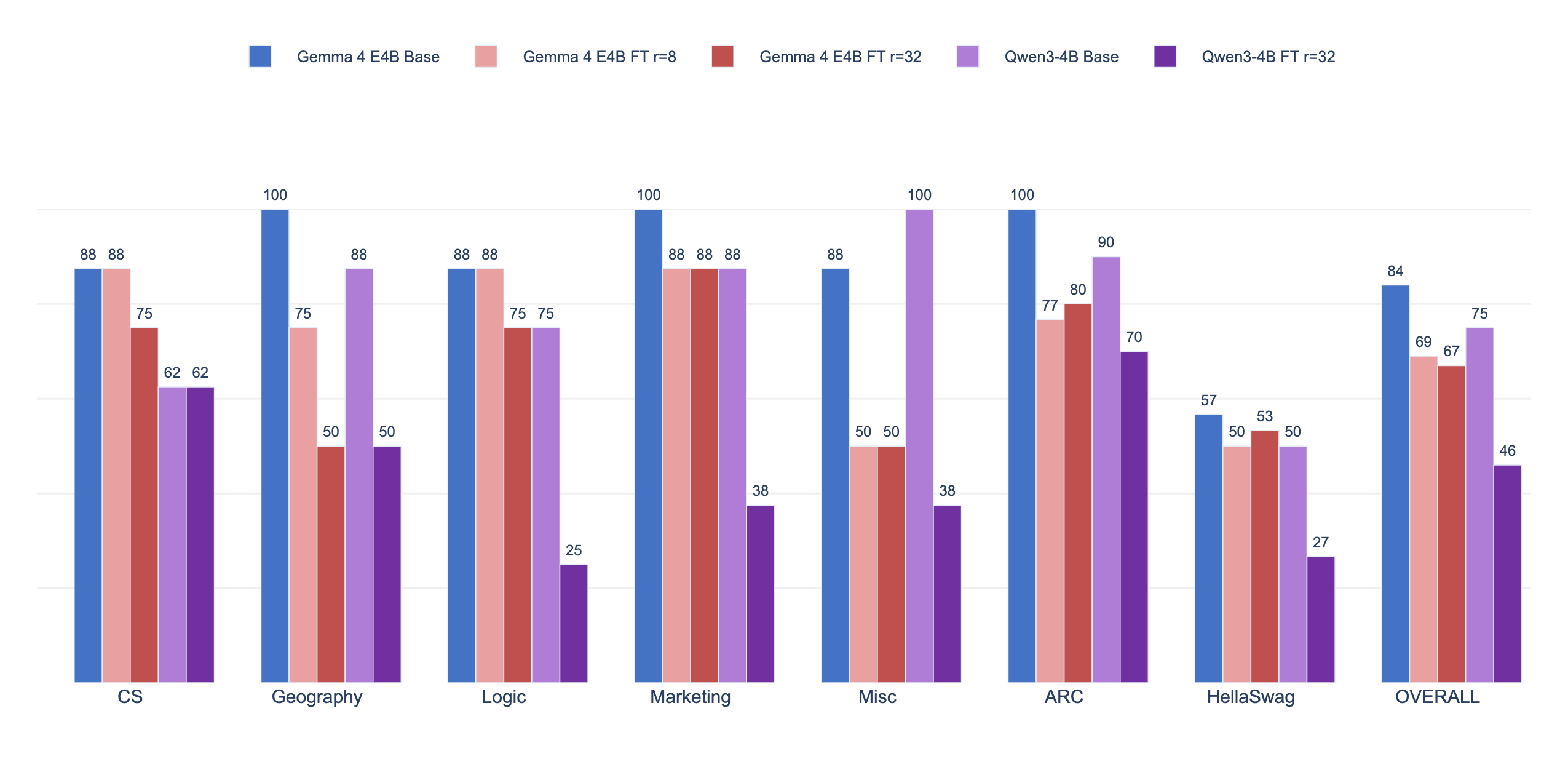}
\caption{Per-benchmark MCQ accuracy for base and fine-tuned models. Qwen3-4B shows severe degradation on Logic (75\%$\rightarrow$25\%), Marketing (88\%$\rightarrow$38\%), and HellaSwag (50\%$\rightarrow$27\%). Gemma degradation is more uniform across benchmarks.}
\label{fig:retention-detail}
\end{figure}

MMLU subjects are selected for a suitable difficulty range on $\sim$4B models (base accuracy 40--100\%): High School Computer Science, High School Geography, Logical Fallacies, Marketing, and Miscellaneous. We avoid niche subjects where the base model scores near zero, making retention difficult to interpret.

The evaluation protocol is fixed across all models: 100 MCQ examples loaded with a fixed random seed; the model is prompted to think step by step and provide a final answer letter (A/B/C/D); generation is capped at 512 tokens; Gemini judges correctness. Overall retention = FT accuracy / base accuracy.

\FloatBarrier

\section{Prompt Example}
\label{app:prompt-example}

We summarize the informed and description-free prompts used for AssetOpsBench scenario~114.

\textbf{Question:} ``What are the failure modes of Chiller 6 that can be identified by analyzing the data from the available sensors?''

\paragraph{Informed prompt.}
The informed prompt contains four components:
\begin{enumerate}
    \item a system instruction describing the model as a planning assistant for industrial asset operations;
    \item the full tool catalog, including MCP server names, tool signatures, argument schemas, and natural-language descriptions;
    \item the required structured output format:
    \texttt{\#Task}, \texttt{\#Agent}, \texttt{\#Tool}, \texttt{\#Args},
    \texttt{\#Dependency}, and \texttt{\#ExpectedOutput};
    \item the user question.
\end{enumerate}
This prompt is approximately 2,400 tokens, of which roughly 2,200 tokens come from the serialized tool catalog.

\paragraph{Gold plan.}

The gold plan contains four tool calls:
\begin{enumerate}
    \item call \texttt{IoTAgent.assets} with \texttt{\{"site\_name": "MAIN"\}} to identify the asset ID for ``Chiller 6'';
    \item call \texttt{IoTAgent.sensors} with the site name and asset ID to retrieve available sensors;
    \item call \texttt{FMSRAgent.get\_failure\_modes} with \texttt{\{"asset\_name": "Chiller 6"\}} to retrieve known failure modes;
    \item call \texttt{FMSRAgent.get\_failure\_mode\_sensor\_mapping} using the failure modes and sensors from the previous steps to determine which failures can be detected by which sensors.
\end{enumerate}

\paragraph{Description-free prompt.}
In the description-free setting, the entire tool catalog is removed. The model receives only the system instruction, output-format specification, and user question, reducing the input to approximately 128 tokens. The fine-tuned model must therefore recover the same tool sequence from internalized tool knowledge rather than from prompt-time tool descriptions.

\FloatBarrier

\end{document}